# Detection and depth estimation for domestic waste in outdoor environments by sensors fusion

Ignacio de L. Páez-Ubieta * Edison Velasco-Sánchez *
Santiago T. Puente * Francisco A. Candelas *

\* AUtomatics, RObotics and Artificial Vision Lab, Computer Science Research Institute, University of Alicante, San Vicente del Raspeig University Campus s/n, 03690, San Vicente del Raspeig, Alicante, Spain (e-mail: {ignacio.paez, edison.velasco, santiago.puente, francisco.candelas}@ua.es).

**Abstract:** In this work, we estimate the depth in which domestic waste are located in space from a mobile robot in outdoor scenarios. As we are doing this calculus on a broad range of space (0.3 - 6.0 m), we use RGB-D camera and LiDAR fusion. With this aim and range, we compare several methods such as average, nearest, median and center point, applied to those which are inside a reduced or non-reduced Bounding Box (BB). These BB are obtained from segmentation and detection methods which are representative of these techniques like Yolact, SOLO, You Only Look Once (YOLO)v5, YOLOv6 and YOLOv7. Results shown that, applying a detection method with the average technique and a reduction of BB of 40%, returns the same output as segmenting the object and applying the average method. Indeed, the detection method is faster and lighter in comparison with the segmentation one. The committed median error in the conducted experiments was $0.0298 \pm 0.0544$ m.

## 1. INTRODUCTION

Numerous efforts have been made in recent years to try to reduce the amount of waste produced in the European Union (EU). According to Eurostat (2020), at the end of 2020, 205.1 million tonnes of domestic waste were produced. That is to say 465.31 kg per person per year. This problem must be taken into account and solved in order to achieve a better future for society. Using robotic and artificial intelligence systems that allow domestic waste to be categorized and manipulated may be a possible solution. This is achieved by measuring the robot-object distances to facilitate the robot's approach maneuvers to the domestic waste for further manipulation, although these parts are outside of this work.

For this purpose, depth has to be calculated. We can find several methods in the literature, such as in Liao et al. (2017), Solak and Bolat (2018) and Sathyamoorthy et al. (2021). In Liao et al. (2017) depth between 0.1 and 2.1 m is calculated using monocular images and scarce 2D laser range data. Concretely, a residual of residual Neural Network (NN) combines category and regression losses for a continuous depth estimation. An error of 0.00442 m is produced. In Solak and Bolat (2018) depth between 0.4 and 1.8 m is calculated using a new hybrid stereovision-based distance-estimation approach. After triangulating, a look up table and a curve-fitting method are used. The observed error was of 0.0266 m. In Sathyamoorthy et al. (2021) it is determined whether two people maintain a safe distance to avoid contagion from Coronavirus disease (COVID-19). To do this, the detection of humans is done using YOLOv3 and, subsequently, the shortest distance is obtained as the depth. The error is 0.09 m for objects between 0.5 and 4.0 m.

Once the technology to use is decided, several techniques are used in Tornero et al. (2022), Nguyen et al. (2021), Sathyamoorthy et al. (2021) and Hernandez-Vicen et al. (2021) to get the depth in which an object is and its size. In Tornero et al. (2022) they use a RGB-D camera from which they get the depth image from the central point of the object's BB. It gives an average error of $0.14 \pm 0.07$ m in the X axis and $0.11 \pm 0.06$ in the Y axis in a range of distances between 0.3 - 3.0 m. In Nguyen et al. (2021) the noisy and background points are eliminated using a clustering method and, after that, depth is calculated as the mean value of the remaining pixels. However, this method works at distances between 0.40 and 1.40 m, with an error of no more than 0.028 m. In Sathyamoorthy et al. (2021) after detecting the people, the image inside the BB is obtained and the average of the 10% of the pixels with the shortest distance is used as depth. In Hernandez-Vicen

* Research work was funded by the Valencian Regional Government and FEDER through the PROMETEO/2021/075 project and the Spanish Government through the Formación del Personal Investigador [Research Staff Formation (FPI)] under Grant PRE2019-088069. The computer facilities were provided through the IDIFEFER/2020/003 project.

et al. (2021) they use the depth estimation to compute the size of an object in order to grasp it.

In this work, we propose a method for detecting domestic waste in outdoor environments, by means of several approaches which use the detected BB or the object's segmentation employing deep learning. Once detected or segmented, with the aim of calculating the distance at which it is located, we use several techniques such as the shorter, median or average value of the pixels are employed in order to get at which depth objects are. Moreover, we aim to increase the detection range up to 6.0 m while maintaining an error similar to the state-of-the-art work.

## 2. METHODOLOGY

### 2.1 Object Detection and Segmentation Methods

For finding the region of the image in which the object is, several NN are used. Two kinds of NN have been considered in this work, since a bunch of them segment the objects, meanwhile other group just detect the BB. The first kind is semantic segmentation with Yolact (Bolya et al. (2019)), SOLO (Wang et al. (2020a)) and SOLOv2 (Wang et al. (2020b)). The second one is the semantic detection with YOLOv5 (Jocher et al. (2022)), YOLOv6 (Li et al. (2022)) and YOLOv7 (Wang et al. (2022)).

These NN were trained to detect domestic waste with our own dataset with a NVIDIA DGX-A100 Tensor Core GPU with 40 GB of RAM memory. The result is classifying or segmenting the objects in one of the following categories: plastic, carton, glass and metal. The dataset is composed of 6504 images in 11 different scenarios, having the same number of samples for each category. We split it in train, validation and test following the 70/20/10 proportion. Results of the training process can be seen in Table 1 and 2. Although having categories during the training, the aim of the current research was getting the depth in which objects are. That is the reason we omit the category part and just detecting domestic waste with no categories. The furthest labeled object is at approximately 6.0 m, so that is the reason behind limiting the depth position of the objects in this work.

Table 1. Semantic segmentation NN comparison.

| Neural network | mAP @0.75 | mAP @0.90 | Recall | Inference time (ms) |
|---|---|---|---|---|
| Yolact [a] | 0.9749 | 0.6416 | 0.8880 | 17.64 |
| Yolact [b] | **0.9847** | **0.7105** | **0.9020** | **17.55** |
| SOLO [a] | 0.9562 | 0.6230 | 0.8811 | 38.75 |
| SOLO light [a] | 0.9271 | 0.5650 | 0.8583 | 30.74 |
| SOLOv2 [c] | 0.5450 | 0.2040 | 0.7950 | 22.38 |
| SOLOv2 light [a] | 0.9578 | 0.6170 | 0.8618 | 25.60 |

Note: [a] ResNet 50, [b] DarkNet 53, [c] ResNet 34

We choose the best two models in detection and segmentation tasks, considering those which have the highest Mean Average Precision (mAP) and recall, and the lowest inference time and number of parameters. Our choice is clear in the segmentation task (Yolact with DarkNet-53) but not in the detection one. Because that, we picked YOLOv5-nano and YOLOv5-small since they offer us a high mAP with a small dataset (6504 images), low number of parameters and low inference time. Considering that, YOLOv5-large, YOLOv6-large and YOLOv7 are discarded. These NN are slower and need more computational resources compared to YOLOv5-nano and YOLOv5-small.

Table 2. Semantic detection NN comparison.

| Neural network | mAP @0.75 | mAP @0.90 | Recall | Parameters (Million) | Inference time(ms) |
|---|---|---|---|---|---|
| YOLOv5-nano | 0.9070 | 0.6538 | 0.8370 | **1.9** | **0.64** |
| YOLOv5-small | **0.9917** | 0.8371 | **0.9920** | 7.2 | 7.410 |
| YOLOv5-large | 0.9918 | 0.8747 | 0.9970 | 46.5 | 14.65 |
| YOLOv6-nano | 0.9630 | 0.7571 | 0.8970 | 4.3 | 7.96 |
| YOLOv6-tiny | 0.9740 | 0.7114 | 0.9070 | 15.0 | 8.26 |
| YOLOv6-large | 0.9790 | 0.8430 | 0.9320 | 58.5 | 14.65 |
| YOLOv7 | 0.8234 | 0.5936 | 0.8100 | 36.9 | 11.10 |

### 2.2 LiDAR-Camera Fusion

Subsequent to the detection of an object in an outdoor environment with the models of detection and segmentation selected in the previous Section 2.1, we implement an object depth estimation method using the fusion of an RGB-D camera and a LiDAR. A calibration of both sensors is required to use the LiDAR data in conjunction with the image from the RGB-D camera. Dhall et al. (2017) presents a LiDAR-Camera calibration method that determines the extrinsic parameters ($^c\mathbf{R}_l$, $^c\mathbf{t}_l$) between the sensors, which represent the rotation and translation matrix between LiDAR and camera. These values are used to convert the LiDAR point cloud to projected points on the image plane. Once the extrinsic parameters are obtained, we calculate the projections of the point cloud ($X_l, Y_l, Z_l$) with (1), where $\mathbf{u}_l$ and $\mathbf{v}_l$ are the coordinates in pixels on the image plane for width and height respectively and $\odot$ represent the Hadamard product between the $(u,v)$ and $e$ obtained from (2), where $\mathbf{M}_c$ is the camera's intrinsic parameters matrix. In addition, $(\mathbf{u}_l, \mathbf{v}_l) \in \mathbb{Z}^+$, and are within the range of the camera's RGB-D image dimensions.

$$\begin{aligned} \mathbf{u}_l &= \boldsymbol{u} \odot (\boldsymbol{e})^{-1} \\ \mathbf{v}_l &= \boldsymbol{v} \odot (\boldsymbol{e})^{-1} \end{aligned} \quad (1)$$

$$\begin{bmatrix} u \\ v \\ e \end{bmatrix} = \mathbf{M}_c \begin{bmatrix} ^c\mathbf{R}_l & ^c\mathbf{t}_l \\ \mathbf{0} & 1 \end{bmatrix} \cdot \begin{bmatrix} X_l \\ Y_l \\ Z_l \\ 1 \end{bmatrix} \quad (2)$$

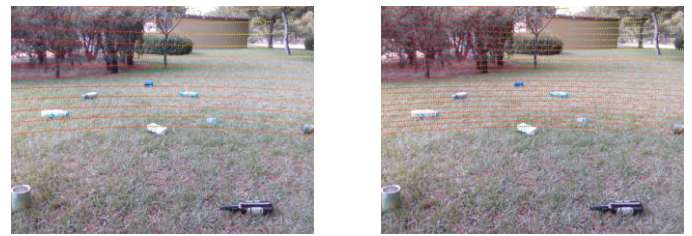

(a) Point cloud on image  (b) Point cloud interpolated

Fig. 1. LiDAR projected points with both normal and interpolated state.

Fig. 1a shows the LiDAR projected points on the image plane. Due to the low vertical density of the point cloud, we interpolated the LiDAR data using Velasco (2022) method, thus increasing the projected points to double, as shown in

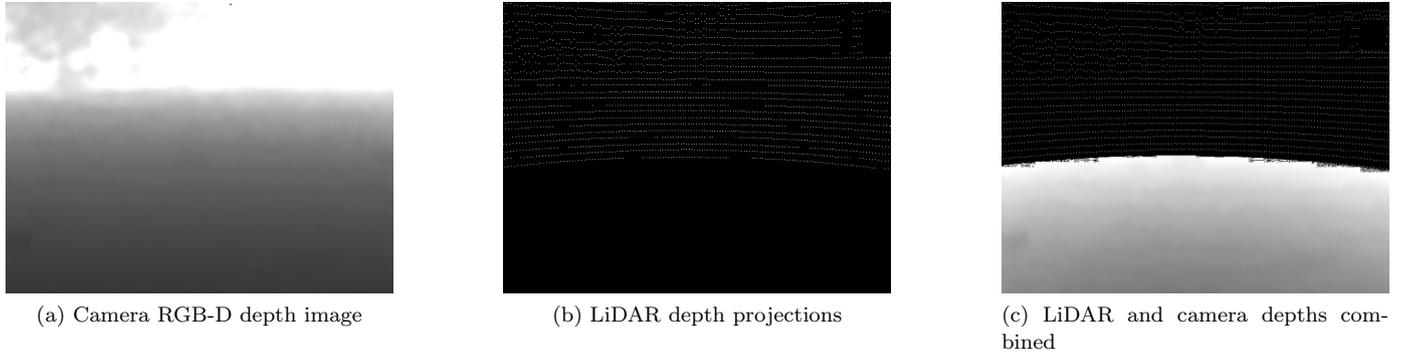

(a) Camera RGB-D depth image　　(b) LiDAR depth projections　　(c) LiDAR and camera depths combined

Fig. 2. RGB-D camera depth image and LiDAR data projection combined on the image plane to determine object distances in outdoor environments.

Fig. 1b. In this way, the point cloud is virtually increased and distances to small objects are estimated, where the original point cloud was not projected. Once the RGB-D camera and LiDAR are calibrated, the camera depth image (see Fig. 2a) and the LiDAR depth data (see Fig. 2b) projected on the image plane are combined, as shown in Fig. 2c. In this way, the area where the LiDAR point cloud was not projected is completed with the RGB-D camera depth image. Thus, the distances of the scenario are obtained, where the RGB-D camera depth image is used for small distances of 0.3 - 3.0 m and the LiDAR projected depth image is used for distances greater than 3.0 m.

Therefore, we estimate the distance between camera and detected object by means of a depth image $\mathbf{dI}$, which has equal dimensions to the RGB image where the object was detected.

### 2.3 Depth Estimation Methods

Regarding the methods to calculate the depth, four are considered. We extract the data from the points in the depth image $\mathbf{dI}$ with the area of objects' BB or mask and get the average (3), median (4), closest (6) and center point (7). Data in the depth image where $\mathbf{dI}(u,v) = 0$ are disregarded, because these are coordinates where the LiDAR point cloud was not reflected or are noise from the deep channel of the RGB-D camera.

$$\text{average} = \frac{1}{n \cdot m} \sum_{u=0}^{n} \sum_{v=0}^{m} [\mathbf{dI}_{bb}(u,v) \neq 0] \quad (3)$$

where $\mathbf{dI}_{bb}$ is the part of the depth image ($\mathbf{dI}$) corresponding to the object detected, $n$ is the height and $m$ is the width of the BB.

$$\text{median} = \begin{cases} \mathbf{dIs}_{\left(\frac{n+1}{2}\right)} & \text{, if } n = \text{odd} \\ \frac{1}{2}\left(\mathbf{dIs}_{\left(\frac{n}{2}\right)} + \mathbf{dIs}_{\left(\frac{n+1}{2}\right)}\right) & \text{, if } n = \text{even} \end{cases} \quad (4)$$

$$\mathbf{dIs} = \text{sort}(\mathbf{dI}_{bb}) \quad (5)$$

where $\mathbf{dIs}$ is the sorted vector of the BB depth values (5).

$$\text{nearest} = \text{argmin}(\mathbf{dI}_{bb} \neq 0) \quad (6)$$

$$\text{center} = \left\{ \mathbf{dI}_{bb}(u', v'), \quad \forall (u', v') = \left\lfloor \left(\frac{u}{2}, \frac{v}{2}\right) \right\rfloor \right\} \quad (7)$$

We then compare the results obtained with the directly measured depth data in order to determine which method is the most accurate.

## 3. EXPERIMENTS AND RESULTS

### 3.1 Experiment Setup

For doing the experiments, we use several scenarios in outdoor environments as show in Fig. 3. These scenarios are different from those used during the training. This method has been implemented on an intel 2.60GHz 6-core processor and a GPU NVIDIA GeForce GTX 1660 Ti with 6 GB of RAM memory. For taking RGB images, an Intel® RealSense™ D435i camera is used with a resolution of 640x480 at 15 Frames Per Second (FPS). If the objects are further from 3.0 m, a LiDAR Velodyne VLP-16 sensor is employed. This sensor allows a measurement frequency of 20 Hz. All this technology is mounted in our research platform *BLUE: roBot for Localization in Unstructured Environments*, which was developed in Del Pino et al. (2020), as shown in the Fig. 4a. For this application, we use the platform as a static element to employ its sensors in order to get the depth estimation in which objects are for a future navigation and grasping project. In the experiments, the LiDAR-camera matrix transformation $(^c\mathbf{R}_l, {}^c\mathbf{t}_l)$ in (2) is defined as $^c\mathbf{T}_l$, as shown in Fig. 4b. In our specific test environment, the objects closer to 1.7 m cannot be detected since they are occluded by the body of the robot.

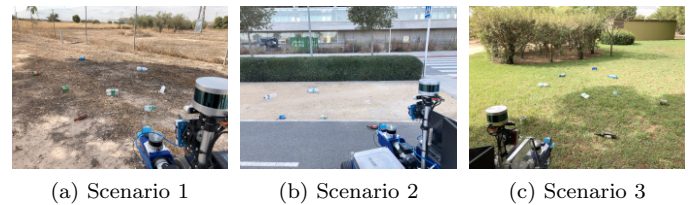

(a) Scenario 1　　(b) Scenario 2　　(c) Scenario 3

Fig. 3. Visualization of the outdoor scenarios with the used sensors and detectable objects positioned.

We conduct three experiments in different scenarios in order to obtain the best method for estimating the depth

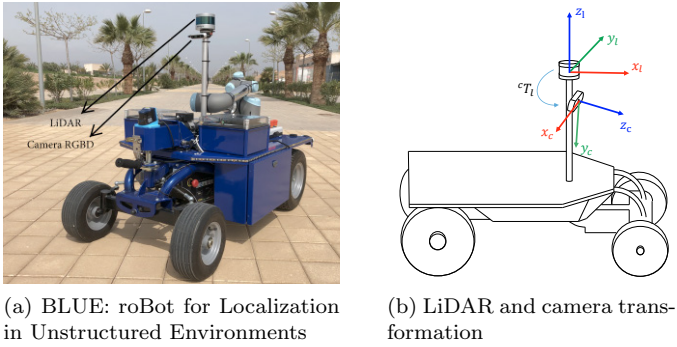

(a) BLUE: roBot for Localization in Unstructured Environments

(b) LiDAR and camera transformation

Fig. 4. Sensors setup used in the experiments.

of objects. In all of them, the ground truth is obtained measuring by hand the distance between the camera and the object, so that we can compare each method with it.

### 3.2 Experiment 1

In this experiment, we checked which method from those described in Section 2.3 obtain the best depth estimation. In addition, we compare whether segmenting the detected object or just using the BB is necessary to get the depth. After obtaining it, we applied a reduction in the BB size to determine if this parameter affects the final results. For that purpose, eight objects are positioned in the scenario and their distances are measured, as shown in Fig. 5a. We then applied Yolact with DarkNet-53 to the image with the methods described in Section 2.3 to segment and obtain a BB for the object.

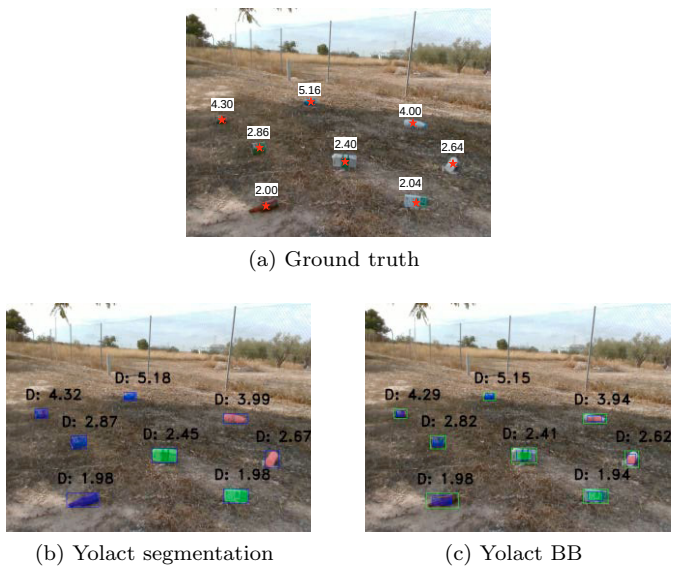

(a) Ground truth

(b) Yolact segmentation

(c) Yolact BB

Fig. 5. First scenario with (a) the ground truth, (b) Yolact segmentation and (c) Yolact detection with BB partial reduction.

The results of this step are shown in Fig. 6. As it can be seen, the lowest error is produced with the average method in Fig. 6a in comparison with the median 6b and the nearest 6c methods. The center method has a huge estimated depth error since it did not overlap with any point of the LiDAR's projection. In that case, the depth value to that point is zero, as shown in Fig. 6d. Applying the average method with Yolact segmentation, Fig. 5b is obtained with an estimated error of $0.0249 \pm 0.0166$ m. Once the best method is identified, we modify the size of the BB, reducing it by a percentage. The reduction of BB makes calculation process lighter since we would have less candidate points to compute in the loop to get the depth.

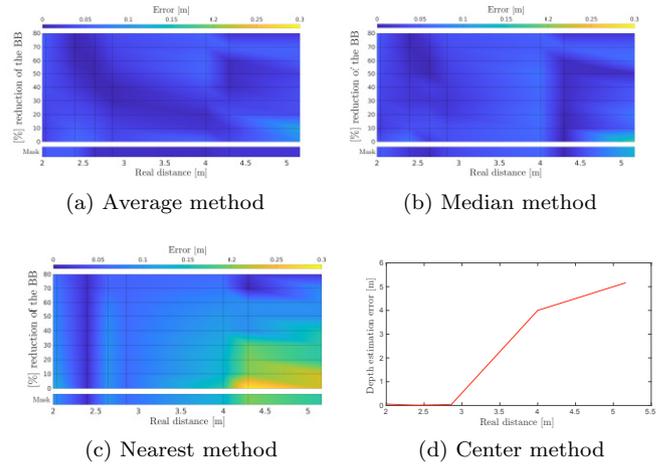

(a) Average method

(b) Median method

(c) Nearest method

(d) Center method

Fig. 6. Error using (a) average, (b) median and (c) nearest methods with several BB reduction. The darker the blue is, the better the depth estimation will be. (d) Depth estimation with center method.

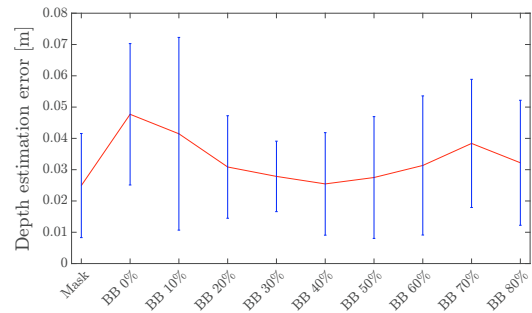

Fig. 7. Mean error and deviation of average method. BB reduction versus accumulated error respectively. Mask and 40% reduction of BB have similar errors.

In Fig. 7 we can see that reducing the size of the BB by 40% returns similar results as if only the pixels of the mask are used. Applying it to Fig. 5b, we obtain Fig. 5c. Depth values in this scenario give us similar results, with an estimated error of $0.02545 \pm 0.01638$ m, which is only 0.55 mm bigger than the error obtained when we used the mask of the object, however the whole process becomes lighter.

### 3.3 Experiment 2

In the second experiment, we tested if using a detection method like YOLO returns a similar error to experiment 1, shown in Section 3.2.

Moving to a new scenario, six objects are positioned and real distances to the objects are measured, obtaining the layout shown in Fig. 8a. We then analyze the image with Yolact with DarkNet-53 and the partially reduced BB method with YOLOv5-small, getting the results shown

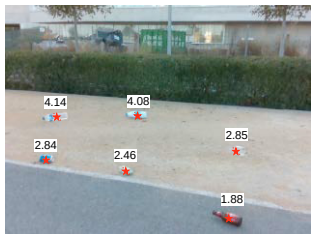

(a) Ground truth

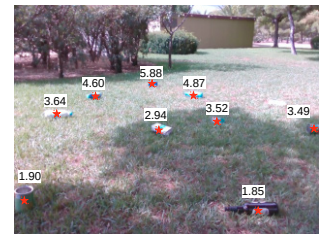

(a) Ground truth

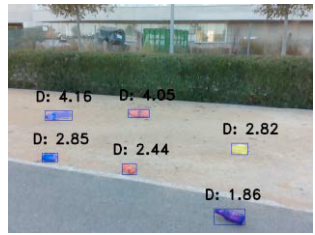

(b) Yolact segmentation

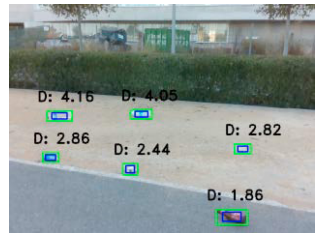

(c) YOLOv5-small detection

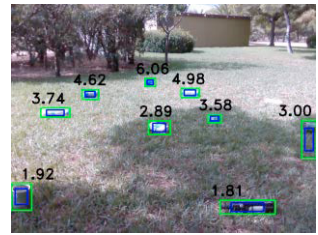

(b) YOLOv5-small

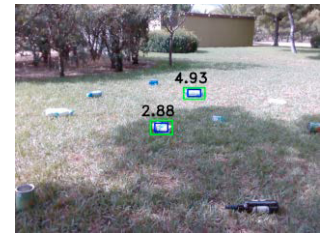

(c) YOLOv5-nano

Fig. 8. Second scenario with (a) the ground truth, (b) Yolact and (c) YOLOv5-small detection with BB partial reduction.

Fig. 10. Third scenario with (a) the ground truth, (b) YOLOv5-small and (c) YOLOv5-nano detection with BB partial reduction.

in Fig. 8b and 8c respectively. In both cases, the average method is used to calculate the depth. To compare which method works better, errors are calculated (see Fig. 9). There is an error of $0.0218 \pm 0.0079$ m in the mask segmentation process, meanwhile in the detection one we have an error of $0.0219 \pm 0.0078$ m.

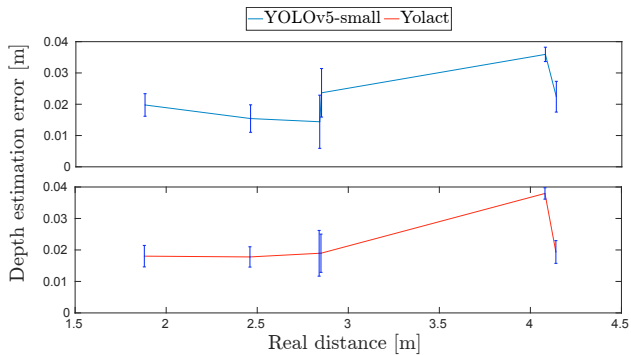

Fig. 9. Error using average method, with both YOLOv5-small and Yolact.

These results point out that segmenting the object is nearly the same as detecting it with the average method and applying a 40% BB reduction, simplifying the estimation of the depth.

### 3.4 Experiment 3

In the third and last experiment, we verified if the detection process could be run faster, losing the least possible detected objects. The 40% reduction in the area of the BB will also be applied.

We put nine objects in a new scenario and, as in previous experiments, we measured the distance by hand in order to build the ground truth, as shown in Fig. 10a. After that, we applied YOLOv5-small and YOLOv5-nano with the BB reduction as in previous experiments, obtaining 10b and 10c respectively. As seen in Table 2, YOLOv5-nano is faster than YOLOv5-small, however we loose track of several objects in the image, making this method unhelpful in detection tasks. Also, we are unable to get an estimated error for YOLOv5-nano since we have only two detections. Although we calculated an error of $0.0839 \pm 0.0771$ m in this scenario for YOLOv5-small, as it can be seen in Fig. 11. There is a clear increment in the error due to the object BB is not properly detected, making the error grow quite a lot.

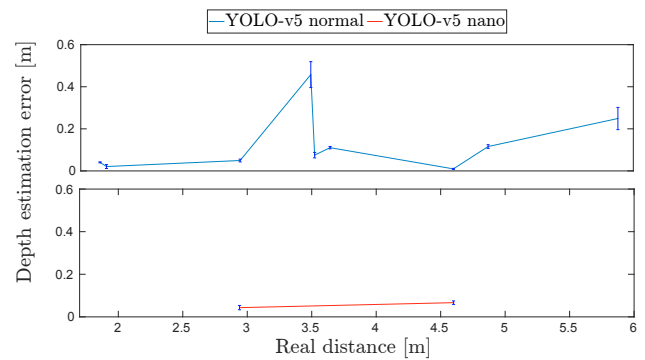

Fig. 11. Error using average method, with both YOLOv5-small and YOLOv5-nano.

## 4. RESULTS AND DISCUSSION

We found that reducing the size of the BB from a detection system like YOLOv5 provides better results than smaller NN or segmentation systems. What is more, apart from having more accurate results, reducing the size of the BB makes the process faster since there are less pixels to process. Our obtained results suggest that a segmentation system, which is a priori a better solution to get the depth of an object works worse than just using a reduced version of the BB. We have also reviewed the state of the art, in which we are able to find other sensors to solve this task. Some examples could be thermal cameras, stereovision or laser. However, we picked the forementioned combination of sensors since they allow us to work in a bigger range of distances. All the state of the art and our method is compared in Table 3. Our method works better than the rest compared in the state of the art, having a bigger range of work with an error bounded in between the other methods, which suggest our method to be fair enough to do the job.

Table 3. State of art and results of technology, methods and committed error in the estimation of depth.

| Method | Range (m) | Technology used | Depth estimator | Error (m) |
|---|---|---|---|---|
| Liao et al. | 0.1 - 2.1 | Monocular image and scarce 2D laser | None | 0.0044 |
| Solak and Bolat | 0.4 - 1.8 | Stereovision-based distance estimation | None | 0.0266 |
| Sathyamoorthy et al. | 0.5 - 4.0 | RGB-D, CCTV and thermal camera | 10% of closest points | 0.0900 |
| Tornero et al. | 0.3 - 3.0 | RGB-D camera | Central point of BB | 0.14 ± 0.07 X<br>0.11 ± 0.06 Y |
| Nguyen et al. | 0.4 - 1.4 | RGB-D camera | Mean value of clustered points | 0.028 |
| Ours | 0.3 - 6.0 | RGB-D camera and LiDAR-camera fusion | Average of 40% reduction of BB | 0.0298 ± 0.0544 |

## 5. CONCLUSION

Thanks to the work presented in this contribution, we estimate the depth at which domestic waste objects are located from a mobile robot in outdoor environments, with a fusion of RGB-D camera and LiDAR. We used several NN for segmentation and detection tasks such as SOLO, Yolact or YOLO to detect the objects. Several techniques were applied in order to obtain the most accurate depth value in comparison with the ground truth. These techniques were median, average or center point. The best method was the average one, followed by the median and nearest ones. Then we reduced the size of the BB in order to see how this parameter affects the detection. We conclude that, using an average method in a 40% reduced BB area by a detection NN like YOLOv5 gives the most accurate depth values compared with the ground truth, having a median error of 0.0298 ± 0.0544 m considering all the performed detections with an execution time of around 20 ms. Some limitations of our process are some BB placed near the margins of the image or misinterpreted with the scenario, making the depth estimation process very difficult. Also, as technology evolves, new NN will arise and take the place of current state of the art methods. The detection time obtained is so low that the method could also be run in real time while the robot navigates. Future works will follow these perspectives in order to continue having results as great are those we currently have.